\documentclass[11pt, letterpaper]{arxiv}
\usepackage[all]{hypcap}
\usepackage[comma,numbers,sort,compress]{natbib}
\usepackage{hyperref}[citecolor=magenta]
\usepackage[capitalize,noabbrev]{cleveref}
\hypersetup{
    colorlinks = true,
    citecolor = {magenta},
}

\usepackage{microtype}

\usepackage{subcaption}
\usepackage{booktabs} %

\usepackage{algorithm}
\usepackage{algorithmic}
\usepackage{mathrsfs}
\usepackage{setspace}

\usepackage{amsmath}
\usepackage{amssymb}
\usepackage{mathtools}
\usepackage{amsthm}
\usepackage[table]{xcolor}
\usepackage{array}
\usepackage{enumitem}
\usepackage{tabularx}
\usepackage{wrapfig}
\usepackage{multirow}
\usepackage[T1]{fontenc}

\usepackage[inkscapelatex=false]{svg}

\setboolean{logo}{true}

\usepackage[most,skins,theorems]{tcolorbox}
\tcbset{
  aibox/.style={
    width=\linewidth,
    top=7pt,
    bottom=2pt,
    colback=blue!6!white,
    colframe=black,
    colbacktitle=black,
    enhanced,
    center,
    attach boxed title to top left={yshift=-0.1in,xshift=0.15in},
    boxed title style={boxrule=0pt,colframe=white,},
  }
}
\newtcolorbox{AIbox}[2][]{aibox,title=#2,#1}

\usepackage{wrapfig}
\definecolor{lightblue}{rgb}{0.22,0.45,0.70}%
\usepackage{tabularx}

\definecolor{rliableolive}{HTML}{BBCC33}
\definecolor{rliableblue}{HTML}{77AADD}
\definecolor{rliablered}{HTML}{EE8866}

\usepackage{crossreftools}
\usepackage[suppress]{color-edits}

\usepackage{dsfont}
\usepackage{nicefrac}
\newtcolorbox{analysisbox}[1][]{
    enhanced jigsaw,
    colback=white,
    colframe=blue!75!black,
    fonttitle=\bfseries,
    boxsep=5pt,
    left=5pt,
    right=5pt,
    top=5pt,
    bottom=5pt,
    title=#1,
}
\definecolor{editInitialResponse}{RGB}{255, 235, 156} %
\definecolor{editBacktrack}{RGB}{0, 0, 139} %
\definecolor{editRevisedResponse}{RGB}{255, 182, 193} %

\usepackage{pifont}
\usepackage{soul}
\definecolor{highlightmistake}{RGB}{255, 179, 179} 
\definecolor{highlightcorrect}{RGB}{179, 255, 179}

\usepackage[capitalize,noabbrev]{cleveref}

\theoremstyle{plain}

\theoremstyle{definition}

\theoremstyle{remark}

\newtcolorbox{solutionbox}{
  colframe=black,
  colback=gray!10,
  boxrule=1pt,
  arc=0pt,
  title=,
  fonttitle=\bfseries
}
\newenvironment{sol}
  {\begin{solutionbox}}
  {\end{solutionbox}}
\newcommand{\BeginSol}{\begin{sol}}
\newcommand{\EndSol}{\end{sol}}

\usepackage[textsize=tiny]{todonotes}

\usepackage{enumitem}

\usepackage{amsmath,amsfonts,bm}

\def\eqref#1{Eq.~\ref{#1}}

\def\1{\bm{1}}

\DeclareMathAlphabet{\mathsfit}{\encodingdefault}{\sfdefault}{m}{sl}
\SetMathAlphabet{\mathsfit}{bold}{\encodingdefault}{\sfdefault}{bx}{n}

\definecolor{codegray}{gray}{0.9}
\definecolor{codepurple}{rgb}{0.58,0,0.82}
\definecolor{codeblue}{rgb}{0.25,0.5,0.5}

\lstdefinelanguage{YAML}{
  morekeywords={selector, sequence, condition, task, no},   %
  keywordstyle=\color{codeblue}\bfseries,                %
  ndkeywords={},                                        %
  sensitive=false,                                       %
  comment=[l]{\#},                                      %
  morecomment=[s]{/*}{*/},                               %
  commentstyle=\color{dkgreen}\ttfamily,               %
  string=[b]",                                          %
  stringstyle=\color{codepurple}\ttfamily,              %
  morestring=[b]',                                         %
  morestring=[b]`,                                         %
  identifierstyle=\ttfamily,                             %
  backgroundcolor=\color{codegray},                      %
  basicstyle=\ttfamily\footnotesize,
  breaklines=true,                                      %
  captionpos=b,                                         %
  frame=single,                                        %
  numbers=left,                                        %
  numberstyle=\tiny\color{gray},                        %
  numbersep=5pt,
  tabsize=2,                                           %
  showspaces=false,                                      %
  showstringspaces=false,                                %
  showtabs=false,                                        %
  xleftmargin=1em,
}
\title{Cogito, Ergo Ludo: An Agent that Learns to Play by Reasoning and Planning}

\author[1,2]{Sai Wang}
\author[2]{Yu Wu}
\author[1]{Zhongwen Xu}

\affil[1]{Tencent}
\affil[2]{Wuhan University}
\correspondingauthor{wuyucs@whu.edu.cn, zhongwenxu@tencent.com}

\begin{document}
\maketitle

\textbf{Abstract:}
The pursuit of artificial agents that can learn to master complex environments has led to remarkable successes, yet prevailing deep reinforcement learning methods often rely on immense experience, encoding their knowledge opaquely within neural network weights. We propose a different paradigm, one in which an agent learns to play by reasoning and planning. We introduce \textit{Cogito, ergo ludo} (CEL), a novel agent architecture that leverages a Large Language Model (LLM) to build an explicit, language-based understanding of its environment's mechanics and its own strategy. Starting from a \textit{tabula rasa} state with no prior knowledge (except action set), CEL operates on a cycle of interaction and reflection. After each episode, the agent analyzes its complete trajectory to perform two concurrent learning processes: \textit{Rule Induction}, where it refines its explicit model of the environment's dynamics, and \textit{Strategy and Playbook Summarization}, where it distills experiences into an actionable strategic playbook. We evaluate CEL on diverse grid-world tasks (i.e., Minesweeper, Frozen Lake, and Sokoban), and show that the CEL agent successfully learns to master these games by autonomously discovering their rules and developing effective policies from sparse rewards. Ablation studies confirm that the iterative process is critical for sustained learning. Our work demonstrates a path toward more general and interpretable agents that not only act effectively but also build a transparent and improving model of their world through explicit reasoning on raw experience.

\begin{section}{Introduction}

The quest to create intelligent agents~\citep{sutton2022quest} capable of mastering complex, interactive environments has been a long-standing goal of artificial intelligence~\citep{rlbook}. Landmark achievements, from Deep Blue's victory in chess to AlphaGo~\citep{alphago,alphagozero,alphazero}'s dominance in Go, have demonstrated the power of computation and search~\citep{bitter_lesson}. More recently, large-scale deep reinforcement learning (RL) has produced agents with superhuman abilities in complex video games \citep{Alphastar, OpenAI_Five}. These systems, however, often learn inefficiently through experience, requiring immense computational resources and encoding their strategic knowledge \emph{implicitly} within the millions of \emph{parameters} of a neural network, rendering their decision-making processes opaque.

The advent of Large Language Models (LLMs) presents a paradigm shift, offering a new foundation for agent design grounded in reasoning and explicit knowledge representation \citep{R1,Gemini}. While early LLM-based agents show promise~\citep{Gemini,understanding_TIR}, they often lack a structured mechanism for continuous learning and adaptation. They may operate in a zero-shot capacity~\citep{hu2025lmgame} or rely on simple memory retrieval~\citep{Gemini}, but they do not fundamentally improve their internal model of the world's mechanics through experience. Similarly, while learned world models \citep{muzero, dreamer, richens2025general} have enabled agents to plan in imagined futures, their models operate on uninterpretable latent states, shrouding their ``understanding'' of the world in a black box. This leaves a critical gap: the need for an agent that not only acts, but truly comprehends its environment in a way that is both effective and interpretable.

In this work, we introduce \textit{Cogito, ergo ludo} (CEL), a novel agent architecture (Figure \ref{fig:introduction}) that learns to master interactive environments not just by acting, but by \textit{reasoning} and \textit{planning}. We propose an agent that leverages an LLM to explicitly reason about its interactions, building and refining a human-readable ``world model''~\citep{wm,rlbook} of its environment and its own strategy from the ground up. Starting from a \textit{tabula rasa} state with no prior knowledge of the game rules, CEL learns purely through a cycle of interaction and reflection, embodying the principle of learning by thinking.

The cornerstone of CEL is its two-phase operational cycle. During an episode, the agent acts decisively by performing a lookahead search with \emph{natural language}, using its current understanding of the world to predict the outcomes of its actions. Crucially, after each episode concludes, the agent enters a \textit{Post-Episode Reflection} phase. In this phase, the LLM analyzes the trajectory of the preceding episode to perform two concurrent learning processes: \textit{Rule Induction}, where it refines its explicit, language-based model of the environment's dynamics; and \textit{Strategy and Playbook Summarization}, where it distills successful and unsuccessful patterns of behavior into an actionable strategic playbook. This refined knowledge base, both the rules of the world and the principles of how to act within it, directly informs the agent's decision-making in subsequent episodes.

We demonstrate the effectiveness of CEL across three distinct grid-world environments: the logical puzzle of Minesweeper, the navigation challenge of Frozen Lake, and the complex planning problem of Sokoban. Our experiments show that CEL successfully learns to master these tasks by autonomously discovering their rules and developing effective strategies. Ablation studies confirm that the iterative, reflective process of refining its internal knowledge is critical to its learning success. Furthermore, we provide qualitative evidence of the architecture's unique interpretability, showcasing the comprehensive, human-readable rulebooks and sophisticated strategic heuristics that it generates \textit{entirely} from raw interaction. Our work presents a step towards agents that not only perform well, but also build a transparent and improving \textit{understanding} of their world.

\begin{figure}[t!]
    \centering
    \includegraphics[width=1\textwidth]{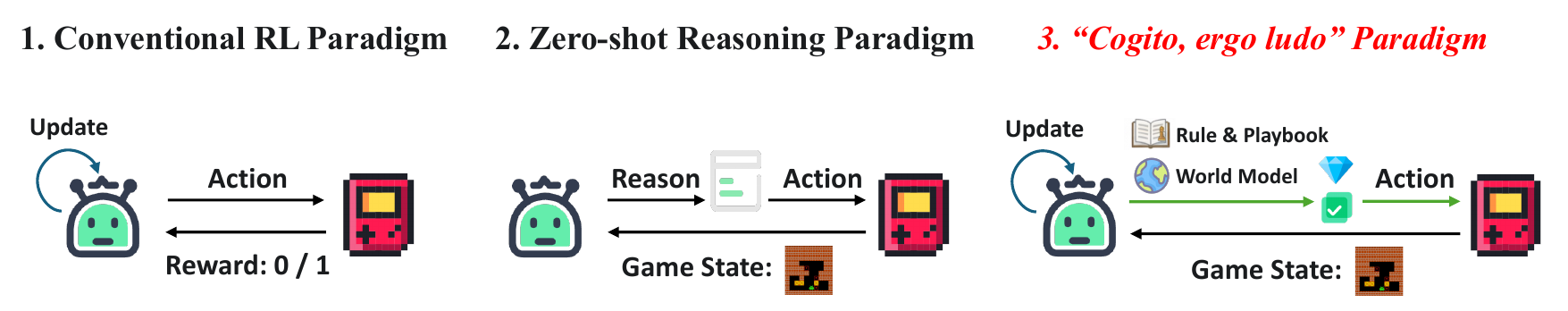}
    \caption{A Comparison of Paradigms for Game-Playing Agents. This figure contrasts three distinct agent architectures: (1) the Conventional RL Paradigm that learns an implicit policy from rewards, by updating policy weights; (2) the Zero-shot Reasoning Paradigm that leverages a static LLM model for decision-making; (3) the \textit{Cogito, ergo ludo} Paradigm, where the agent's policy is trained by RL while a persistent knowledge base (Rule \& Playbook) is built and passed \emph{across episodes.}}
    \label{fig:introduction}
\end{figure}

\end{section}

\begin{section}{Related Work}\label{sec:related}

Our research builds upon decades of work in artificial intelligence, drawing from and extending three key areas: the paradigm of large-scale deep reinforcement learning, the development of learned world models for planning, and the nascent field of agents driven by Large Language Models.

\textbf{The Apex of Deep Reinforcement Learning.}
Landmark achievements such as DeepMind's AlphaStar~\citep{Alphastar} and OpenAI Five~\citep{OpenAI_Five} demonstrated that deep reinforcement learning (RL) could attain superhuman performance in complex real-time strategy games. These systems operate at a massive scale, training for thousands of GPU-years on billions of game frames. Their strategic acumen is implicitly encoded within the weights of enormous neural networks, learned through vast experience. While immensely powerful, this approach is characterized by high sample complexity and the opaque nature of the resulting policies. Our work diverges from this paradigm by pursuing a more sample-efficient and interpretable approach, where knowledge and strategy are explicitly represented in natural language. The AlphaZero algorithm~\citep{alphago,alphagozero,alphazero} uses the power of Monte-Carlo Tree Search (MCTS)~\citep{MCTS} with a deep neural network, achieving superhuman performance in Chess, Shogi, and Go. But crucially, it was provided with a perfect model of the environment -- the game rules, while our proposed architecture accumulates the knowledge of the game rules purely by interaction.

\textbf{Planning with Learned World Models.}
 MuZero~\citep{muzero} learned a latent model to predict future rewards, policies, and values, enabling effective lookahead search without being given the rules. This principle of learning and planning in imagined trajectories has been further advanced by algorithms like Dreamer~\citep{dreamer}, which learns a robust world model that allows it to master a vast suite of diverse domains, from Atari to Minecraft, with a single set of hyperparameters. Our Language-based World Model (LWM) shares this objective of predicting environmental dynamics. However, we draw a critical distinction: whereas the models in MuZero and Dreamer operate on uninterpretable latent states, our LWM is grounded in explicit, human-readable rules and transition dynamics that are themselves inferred from experience. This language symbolic foundation allows the agent to reason about and refine its understanding of the world's mechanics in natural language.

\textbf{Language as the Algorithm vs. Language in the Architecture.}
A distinct approach is Natural Language Reinforcement Learning (NLRL)~\citep{NLRL}, which seeks to fundamentally redefine the core components of RL, such as the value function and Bellman equation, entirely within the domain of natural language. In NLRL, the value of a state is not a scalar but a descriptive text, and the policy improvement step is performed by an LLM reasoning over these linguistic value judgments. While both our approaches leverage LLMs for reasoning, our philosophy and architecture differ significantly. Rather than reformulating the RL algorithm itself into language, our framework treats the LLM as the orchestrator of a cognitive architecture composed of distinct, language-grounded modules. 

\textbf{LLMs as Agent Architectures.}
More recently, the advent of LLMs has catalyzed a new approach to agent design. Frameworks like GEM~\citep{gem} and LMGame-Bench~\citep{hu2025lmgame} provide environments and harnesses to evaluate LLM agents, highlighting challenges in perception, memory, and long-horizon planning. Gemini 2.5 Pro~\citep{Gemini} showcases its success in complete Pokémon game playing, demonstrating the strong zero-shot reasoning abilities of the frontier LLMs. A particularly relevant approach is PORTAL~\citep{xu2025agents}, which uses an LLM as a ``policy architect'' to generate behavior trees in a domain-specific language. Unlike PORTAL, our method uses LLM as the core for planning, acting and accumulating knowledge, where the LLM directly interacts with environments.

Our work builds upon this foundation but proposes a more comprehensive cognitive architecture. Our agent learns and maintains a suite of distinct, yet interconnected, cognitive components: an explicit world model of environmental dynamics, a set of game rules, a strategic playbook, and a language-based value function. The cornerstone of our method is the post-episode reflection phase, where the LLM analyzes interaction trajectories to iteratively and simultaneously refine both its understanding of the world's rules and its own strategic playbook. This creates a cycle of self-improvement that is explicit, interpretable, and broadly applicable to any interactive environment.

\end{section}

\begin{section}{Method}

We model the agent's interaction with its environment as a Markov Decision Process (MDP)~\citep{MDP}, formally defined by the tuple $(\mathcal{S}, \mathcal{A}, \mathcal{P}, \mathcal{R}, \gamma)$. In this framework, $\mathcal{S}$ represents the set of states, $\mathcal{A}$ the set of actions, and $\gamma \in [0, 1]$ the discount factor. The state transition function $\mathcal{P}(s_{t+1}|s_t, a_t)$ specifies the probability of transitioning to state $s_{t+1}$ from state $s_t$ upon taking action $a_t$. The reward function $\mathcal{R}(s_t, a_t)$ yields an immediate reward $r_{t+1}$. The agent's objective is to learn a policy $\pi(a|s)$, that maximizes the expected discounted return, defined as $G_t = \sum_{k=0}^{\infty} \gamma^k r_{t+k+1}$~\citep{rlbook}.

Our central methodological contribution is to employ a single Large Language Model (LLM), denoted $\mathcal{L}$, to instantiate and manage all of the agent's cognitive functions.
Our framework moves beyond the static zero-shot paradigm by continuously training an LLM based on the outcomes of the agent's interactions, allowing it to improve its core reasoning and planning capabilities over time.
We represent all information pertaining to the interaction: states ($s$), actions ($a$), rewards ($r$), inferred environmental dynamics ($\mathcal{G}$), and strategic guidelines ($\Pi$), as natural language strings. The agent's learning unfolds over a series of episodes, indexed by $k$, where each episode consists of discrete time steps, indexed by $t$. The LLM's reasoning process is made explicit through a chain-of-thought~\citep{cot}, which we denote by $C$. By reasoning and planning, the CEL agent $\mathcal{L}$ learns to interact with the environment and maximize its rewards.

\subsection{Language-based World Model}
The LLM functions as a world model~\citep{wm,dreamer}, tasked with predicting the dynamics of the environment. Given the current state $s_t$ and a candidate action $a_t$, the world model forecasts the subsequent state $\hat{s}_{t+1}$ and immediate reward $\hat{r}_{t+1}$. This prediction is conditioned on the agent state $s_t$, a potential action $a_t$, and the agent's current understanding of the environment's rules, $\mathcal{G}_k$. The model first generates a reasoning trace, $C_{WM}$ (for \textbf{W}orld \textbf{M}odel), before outputting its predictions:
\begin{equation}
    (C_{WM}, \hat{s}_{t+1}, \hat{r}_{t+1}) \sim p_{\mathcal{L}}(\cdot | s_t, a_t, \mathcal{G}_k),
\end{equation}
where $p_{\mathcal{L}}$ is the probability distribution over text sequences generated by the LLM. Critically, the outputs $\hat{s}_{t+1}$ and $\hat{r}_{t+1}$ are not structured data types (e.g., a state tensor or a scalar value) but are descriptive natural language strings, as illustrated in Figure~\ref{fig:thinking}.

This predictive capability is the foundation for explicit planning~\citep{rlbook}. By querying the world model for each potential action, the agent can simulate and evaluate a set of possible future outcomes, a process that is central to its decision-making~\citep{muzero,richens2025general}.

\begin{figure}[t!]
    \centering
    \includegraphics[width=1\textwidth]{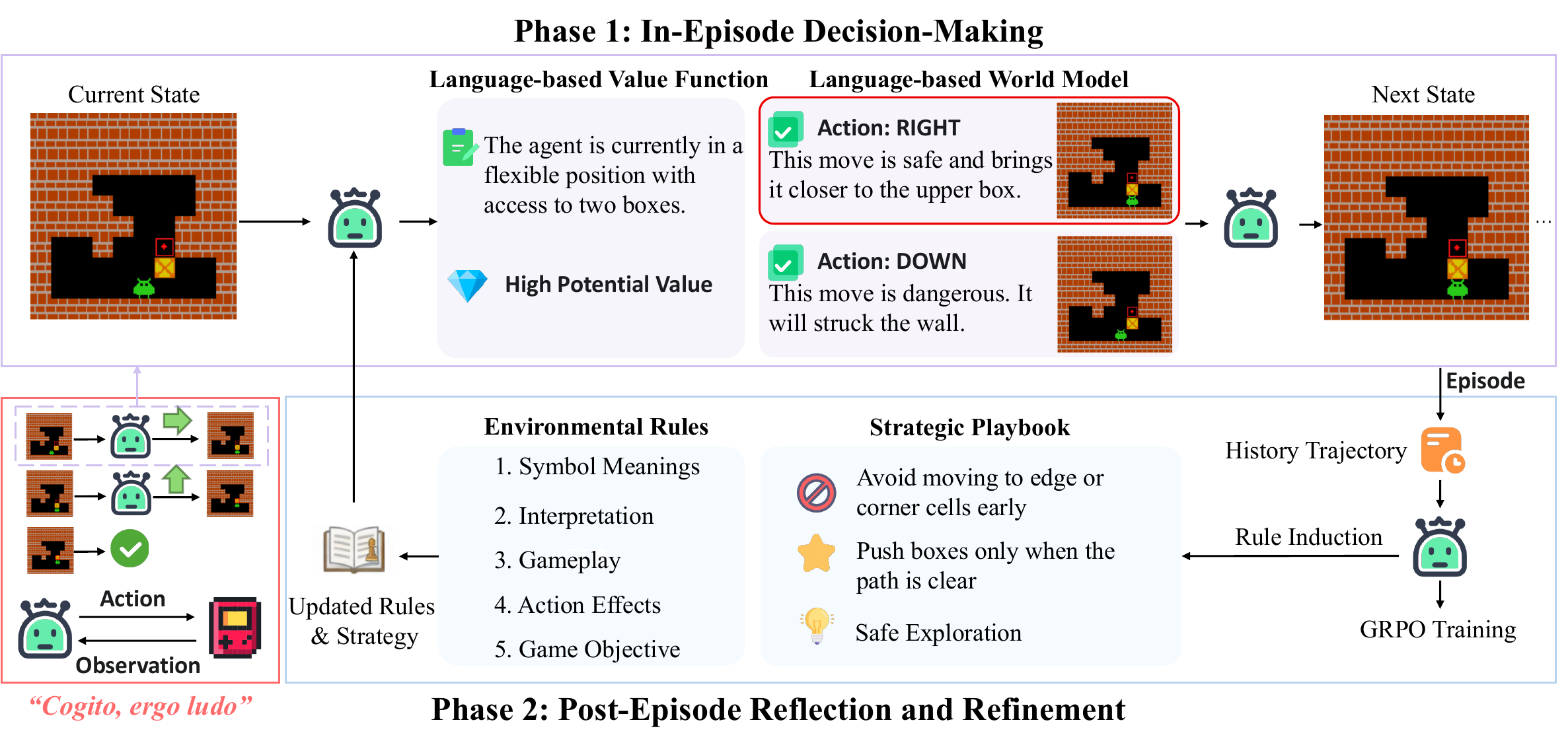}
    \vspace{-12pt}
    \caption{An overview of the \textit{Cogito, ergo ludo} (CEL) agent's two-phase operational cycle. In Phase 1, the agent leverages its Language-based World Model (LWM) to predict the outcomes of potential actions and its Language-based Value Function (LVF) to evaluate the desirability of the resulting states, ultimately selecting the optimal action. In Phase 2, it reflects on the episode's trajectory to update its explicit knowledge base (Environmental Rules and Strategic Playbook). The agent continuously improves through this dual learning loop, which not only refines its explicit knowledge but also trains the LLM's internal parameters based on the final outcome.
    }
    \label{fig:mainfigure}
\end{figure}

\subsection{Induction of Environmental Dynamics}
Following each episode $k-1$, the agent enters a reflective phase to refine its understanding of the environment's mechanics. The LLM performs rule induction by analyzing the trajectory of the concluded episode, $\tau_{k-1}$, in light of its previously held rules, $\mathcal{G}_{k-1}$. A trajectory is the sequence of state-action-reward tuples recorded during the episode:
\begin{equation}
    \tau_{k-1} = \{ (s_0, a_0, r_1), (s_1, a_1, r_2), \dots, (s_{T_{k-1}-1}, a_{T_{k-1}-1}, r_{T_{k-1}}) \}.
\end{equation}
The LLM processes this experiential data to generate an updated, more accurate set of rules $\mathcal{G}_k$:
\begin{equation}
    (C_{\mathcal{G}}, \mathcal{G}_k) \sim p_{\mathcal{L}}(\cdot | \tau_{k-1}, \mathcal{G}_{k-1}),
\end{equation}
where $C_{\mathcal{G}}$ is the reasoning trace for updating the environment's \textbf{G}overning dynamics (or \textbf{G}ame rules in game environments). We assume the agent begins with no prior knowledge (i.e., \textit{tabula rasa}); the initial rule set $\mathcal{G}_0$ is empty, and all subsequent knowledge is derived purely from interaction~\citep{welcome}.

\subsection{Strategy and Playbook Summarization}
In parallel with rule induction, the agent updates its high-level strategy. After episode $k-1$, the LLM synthesizes the trajectory $\tau_{k-1}$ and the final outcome $Z_{k-1}$ (e.g., success/failure, final score) to update a strategic playbook, $\Pi_k$. This process distills successful and unsuccessful patterns of interaction into explicit, actionable advice:
\begin{equation}
    (C_{\Pi}, \Pi_k) \sim p_{\mathcal{L}}(\cdot | \tau_{k-1}, Z_{k-1}, \Pi_{k-1}),
\end{equation}
where $C_{\Pi}$ is the reasoning trace for the \textbf{P}laybook update, and $\Pi_0$ is initialized with a general-purpose prompt.

This mechanism contrasts sharply with conventional reinforcement learning~\citep{dqn,IMPALA}, where strategy is implicitly encoded within the weights of a neural network. In such systems, adaptation is often slow, sample-inefficient (e.g., requiring millions of interaction frames~\citep{dqn,IMPALA}), and opaque. Our approach externalizes strategy into an explicit, interpretable text playbook. Insights from a single episode can be immediately incorporated into the agent's prompt context for the subsequent episode. This mechanism facilitates rapid \emph{in-context learning}, dramatically accelerating strategic adaptation.

\subsection{Language-based Value Function}
To guide its planning, the agent employs the LLM $\mathcal{L}$ as a language-based value function. This component estimates the value of a state $\hat{v}(s_t)$, by providing a qualitative, linguistic assessment of the long-term potential for success from that state. This evaluation is conditioned on both the current environmental rules $\mathcal{G}_k$ and the strategic playbook $\Pi_k$:
\begin{equation}
    (C_V, \hat{v}(s_t)) \sim p_{\mathcal{L}}(\cdot | s_t, \mathcal{G}_k, \Pi_k).
\end{equation}
Here, $C_V$ is the reasoning trace for the \textbf{V}alue estimation.
This function provides the agent with a crucial heuristic by assessing the current state's long-term potential, which is essential for effective planning.

\subsection{The Agent's Operational Cycle}
The agent's operation is structured as a cyclical pipeline that alternates between two phases: in-episode decision-making and post-episode reflection (Figure \ref{fig:mainfigure}). This architecture decouples rapid, step-by-step action selection from a more deliberate, offline knowledge consolidation process, with the LLM orchestrating both.

\textbf{Phase 1: In-Episode Decision-Making.}
During an episode $k$, the agent operates with a fixed set of environmental rules $\mathcal{G}_k$ and a strategic playbook $\Pi_k$. At each time step $t$, it performs a structured reasoning process to select an action. First, the agent's \textit{Language-based Value Function} (LVF) assesses the desirability of the current state, $s_t$, providing a high-level, holistic evaluation of its strategic potential. Concurrently, for each available action $a$, the agent's \textit{Language-based World Model} (LWM) performs a one-step lookahead search to simulate the resulting state $\hat{s}_{t+1}$ and reward $\hat{r}_{t+1}$. The agent then commits to the action that the LWM predicts will lead to the most favorable outcome. The resulting $(s_t, a_t, r_{t+1})$ tuple is then recorded in the episode's trajectory $\tau_k$.

\textbf{Phase 2: Post-Episode Reflection and Refinement.}
Once an episode concludes, the agent enters the reflection phase to update its internal knowledge base. It performs \textit{Rule Induction} by providing the LLM with the complete trajectory $\tau_k$ and the prior rule set $\mathcal{G}_k$ to produce a refined set of rules $\mathcal{G}_{k+1}$. Concurrently, it engages in \textit{Strategy and Playbook Summarization}, where the LLM processes $\tau_k$ and the final outcome $Z_k$ to distill key lessons, updating the playbook to $\Pi_{k+1}$.
The episode's final outcome (e.g., success or failure) provides a reward signal that is used to train the agent's core LLM, making it progressively more effective at planning and strategic reasoning in future episodes.

This two-phase cycle enables the agent to both act decisively on its current understanding and systematically improve its environmental model and strategic acumen.

\end{section}

\begin{section}{Experiments}\label{sec:exps}
\subsection{Game Environments}
We evaluate our method on three classic grid-world environments: Minesweeper, Frozen Lake and Sokoban. These environments serve as common benchmarks for tasks with sparse rewards in reinforcement learning.
Detailed descriptions of the environments are provided in Appendix~\ref{sec:appendix_env}.

In our experimental setup, all three environments are configured with a sparse reward signal. The agent receives a reward only at the conclusion of the game, receiving +1 for successfully completing the objective and 0 otherwise. Crucially, we DO NOT provide the agent with any explicit game rules. It must learn the dynamics of each environment solely through interaction, with its knowledge limited to the set of available actions. This setup, combining sparse rewards with unknown rules, presents a significant reasoning and planning challenge.

\subsection{Implementation Details}

We conducted experiments using rLLM~\citep{rllm2025}, backed by verl~\citep{sheng2024hybridflow}.
We use the Qwen3-4B-Instruct~\citep{qwen3} model to interact with the environments.
We evaluate the performance over 32 randomly sampled seeds for each game. For each of these seeds, we conduct 8 independent trials, and report the average success rate over the total 256 playthroughs per game.
The rule update frequency is set to once every 5 episodes. We use GRPO~\citep{GRPO,R1} for LLM post-training and set the maximum response length to 8,192 tokens to encourage the model to think, reason, and plan. The outcome reward is used for optimizing the LLM.

\subsection{Results}

\begin{figure}[t!]
    \centering
    \includegraphics[width=1\textwidth]
    {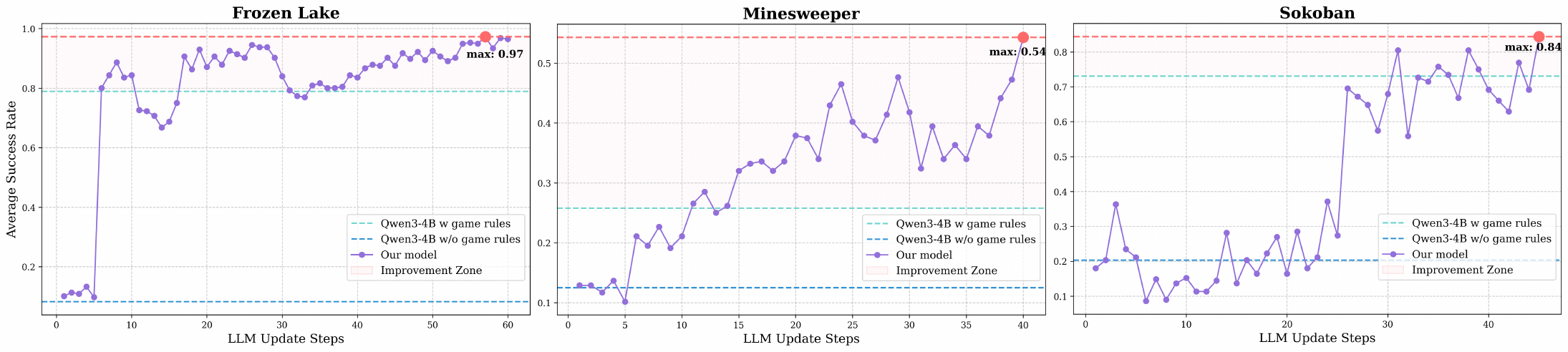}
    \caption{Learning curves of our agent on FrozenLake (left), Minesweeper (center), and Sokoban (right). The plots show the average success rate (y-axis) plotted against the number of LLM update steps (x-axis).
    Starting without any explicit rules, the agent's consistent improvement across these diverse tasks showcases the effectiveness of its autonomous rule discovery and policy learning.}
    \label{fig:train_curves}
\end{figure}

Figure \ref{fig:train_curves} illustrates our CEL agent's learning performance across the three environments.
We compare CEL against two zero-shot baselines operating with and without the ground-truth game rules, respectively.
Despite starting with no explicit game rules, the agent demonstrates a clear and positive learning trend in all tasks, validating the effectiveness of our interaction-reflection cycle. In the logical puzzle of Minesweeper, the agent exhibits steady improvement, with its success rate progressively climbing to a peak of 54\%.
Notably, this surpasses the 26\% success rate of the baseline agent that was explicitly provided with the ground-truth game rules, suggesting that our method of autonomous rule discovery and strategy refinement leads to a more effective policy. 
A different learning dynamic emerged in the complex planning puzzle of Sokoban, where the agent's performance showed a distinct ``breakthrough'' pattern, increasing sharply to an 84\% success rate after an initial period of exploration. This highlights its ability to uncover critical insights for solving multi-step problems. 
The agent's efficiency was most apparent in the Frozen Lake navigation task, where it learned with remarkable speed to achieve a near-perfect success rate of 97\% within the first 10 episodes. Collectively, these results showcase the general applicability and effectiveness of our approach, as it successfully masters diverse tasks ranging from logical deduction to long-horizon planning by autonomously discovering environmental rules and iteratively refining its own strategy from raw interaction.

\subsection{Ablation Study}
\begin{wrapfigure}{r}{0.5\textwidth}
    \vspace{-12pt} %
    \centering
    \includegraphics[width=\linewidth]{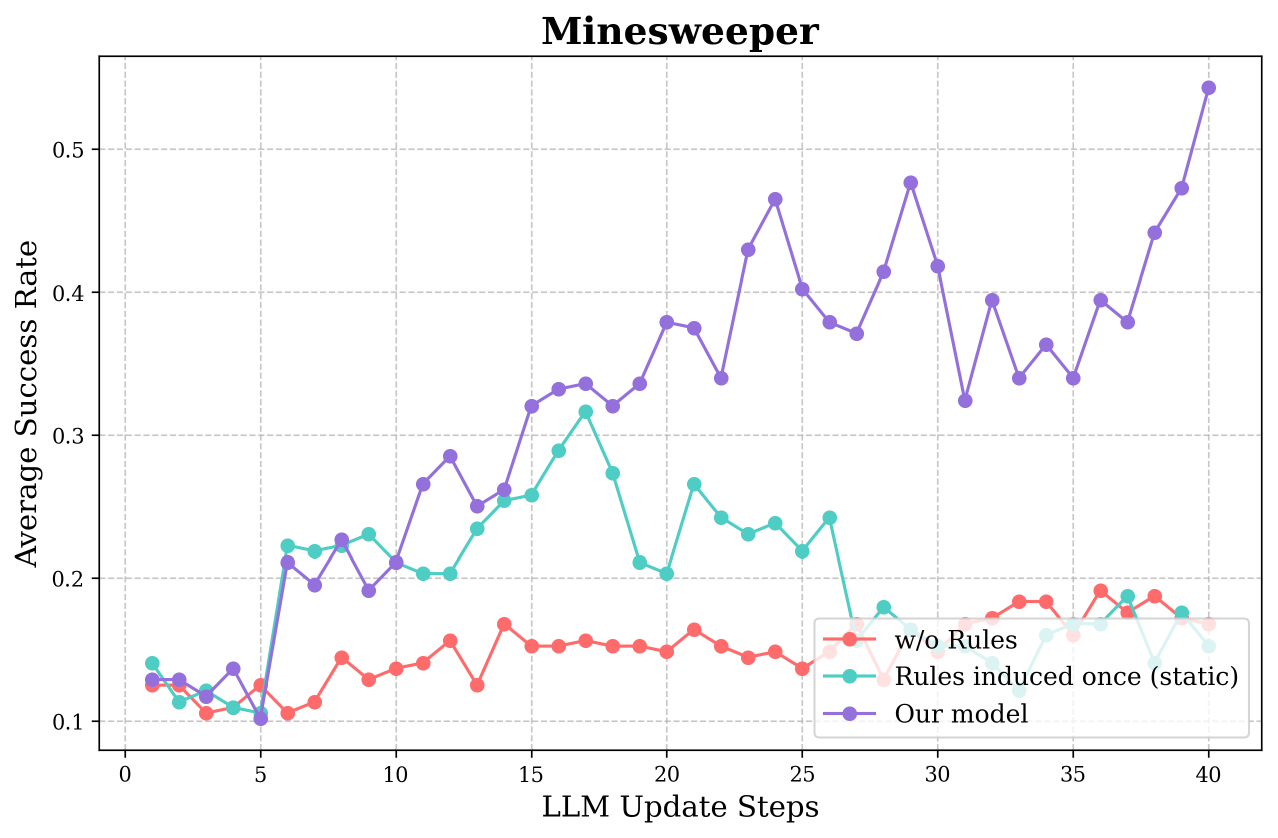}
    \vspace{-12pt}
    \caption{Ablation study illustrating the critical role of iterative \textbf{Induction of Environmental Dynamics}. The full model (blue) significantly outperforms variants with static rules (teal) or no rules (red), demonstrating that continuous refinement of the learned rulebook $\mathcal{G}_k$ is essential for sustained performance improvement.}
    \label{fig:minesweeper_ablation}
    \vspace{-10pt} %
\end{wrapfigure}
We conducted an ablation study in the Minesweeper environment to test the necessity the component of the iterative Induction of Environmental Dynamics, with results shown in Figure~\ref{fig:minesweeper_ablation}.
The baseline agent, operating without the Rule Induction mechanism (``w/o Rules''), exhibits a largely flat learning curve, with its success rate stagnating at a low level.
This confirms that the ability to infer and utilize a model of the environment's dynamics is fundamental to achieving competence. A second variant, which performs Rule Induction only once and then uses a static rule set (``Rules induced once''), shows initial improvement but quickly stagnates and its performance degrades, suggesting its initial rules were incomplete or inaccurate. In stark contrast, our full CEL agent, which engages in the post-episode reflection and refinement phase to continuously update its rule set $\mathcal{G}_k$, shows a robust and sustained learning trajectory, significantly outperforming both ablated versions. This comparison unequivocally demonstrates that the iterative refinement of the agent's world model is a critical component of our architecture.

\subsection{Case Study}

\subsubsection{In-Episode Decision-Making}

Figure~\ref{fig:thinking} provides a qualitative snapshot of the agent's In-Episode Decision-Making process. The examples showcase how the agent performs a one-step lookahead search, a process relying on the synergy between its core cognitive components. First, the agent employs its Language-based Value Function (LVF) to produce a holistic, linguistic assessment of the current state's potential, $\hat{v}(s_t)$. In the Minesweeper example, it correctly identifies the state as having ``high strategic value''. Next, for each viable action, the agent utilizes its Language-based World Model (LWM), conditioned on its learned rules $\mathcal{G}_k$, to simulate the immediate future, predicting the next state $\hat{s}_{t+1}$ and reward $\hat{r}_{t+1}$. It accurately forecasts that the ``(0, 3)'' action in Minesweeper solves the puzzle. By comparing the predicted outcomes, the agent selects the action leading to the most favorable consequence, highlighting how its explicit, language-based reasoning drives intelligent planning.

\subsubsection{Autonomous Rule Discovery}

Figure~\ref{fig:rule} presents an example of the agent's learned rulebook $\mathcal{G}_k$ for Minesweeper, a direct output of the Induction of Environmental Dynamics process. Synthesized from its interaction trajectory and starting from a \textit{tabula rasa} rule, the generated rules are remarkably comprehensive and accurate, covering everything from \textit{Symbol Meanings} to the ultimate \textit{Game Objective}. As defined in our method, this explicit, human-readable rule set $\mathcal{G}_k$ is the critical information that grounds the agent's cognitive functions. It provides the foundation for the Language-based World Model to predict future states and for the Language-based Value Function to estimate state values, thereby enabling all subsequent planning.

\begin{figure}[!t]
    \centering
    \includegraphics[width=1\textwidth]{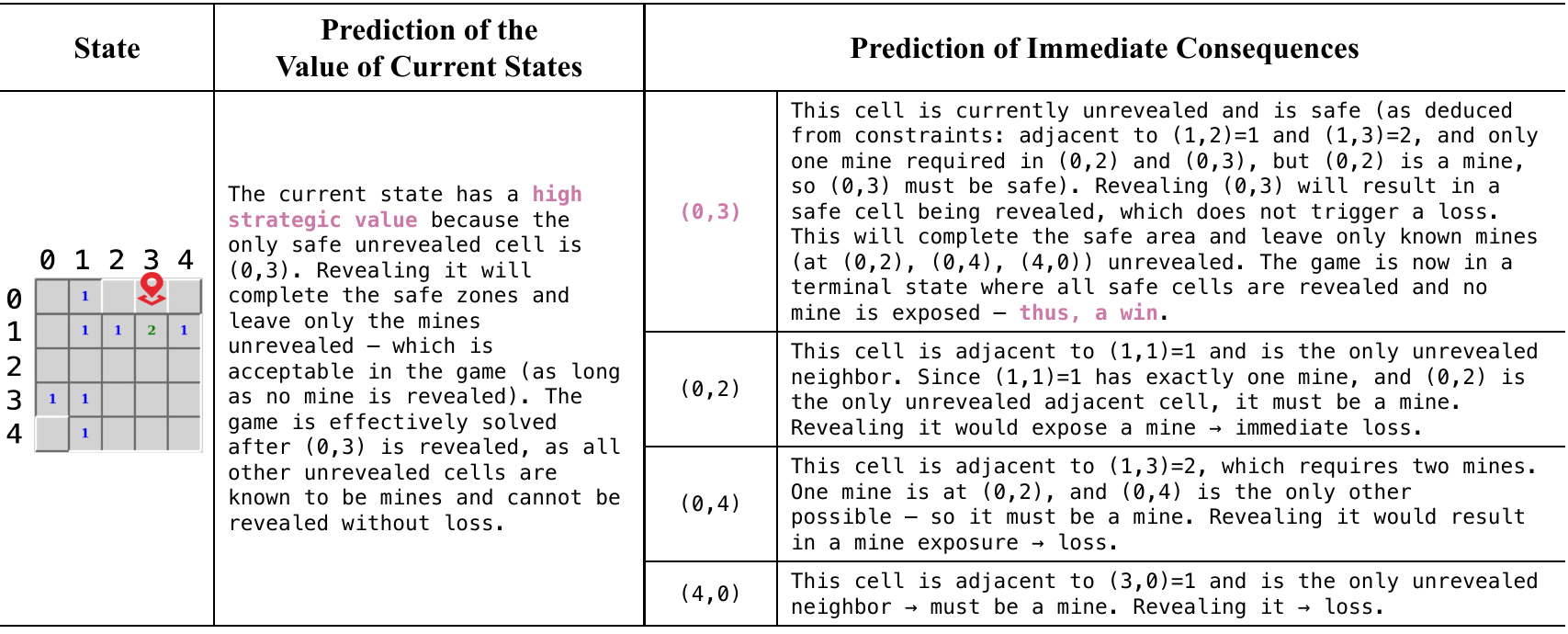}
    \vspace{-6pt}
    \caption{The agent's In-Episode Decision-Making process. At each step, the agent uses its Language-based Value Function (LVF) to assess the current state's value (middle column). It then employs its Language-based World Model (LWM) to predict the consequences of each action (right column). The agent selects the action leading to the outcome with the highest predict value. 
    }
    \label{fig:thinking}
\end{figure}

\begin{figure}[t!]
    \centering
    \includegraphics[width=1\textwidth]{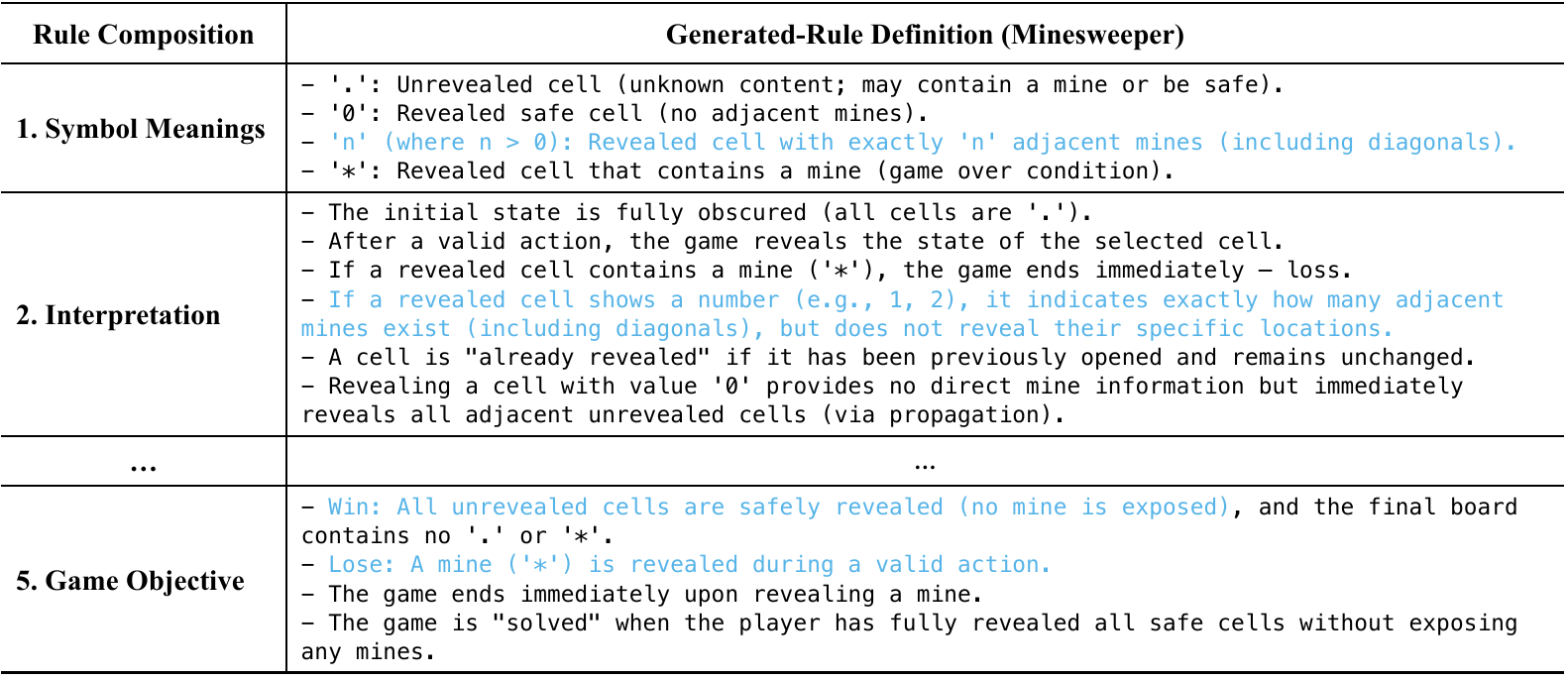}
    \vspace{-6pt}
    \caption{An illustrative excerpt of the agent's learned rulebook $\mathcal{G}_k$ for Minesweeper, generated via the Induction of Environmental Dynamics process. Starting from no prior knowledge, the agent synthesizes a comprehensive and accurate set of rules from its interaction trajectory.
    Please refer to Figure~\ref{fig:minesweeper_rule_full} in the Appendix for the complete rulebook.
    }
    \label{fig:rule}
    \vspace{-8pt}
\end{figure}

\subsubsection{Emergent Strategy and Playbook Generation}

In parallel with rule induction, our agent constructs a strategic playbook $\Pi_k$, via the Strategy and Playbook Summarization process, synthesized in Figure~\ref{fig:strategy}. As defined, the LLM analyzes an episode's trajectory $\tau_{k-1}$ and outcome $Z_{k-1}$ to distill experiences into actionable advice. The emergent knowledge exhibits a sophisticated hierarchy, from tactical \textit{Methods} like Constraint Propagation to high-level \textit{Principles} like Safe Exploration. The discovery of these expert-level heuristics from raw interaction highlights the agent's capacity for strategic abstraction. This explicit playbook $\Pi_k$ is then used alongside the rule set $\mathcal{G}_k$ in the next episode to condition the Language-based Value Function, enabling more nuanced, strategically-aware judgments and forming a direct feedback loop from experience to adaptation.

\begin{figure}[t!]
    \centering
    \includegraphics[width=0.8\textwidth]{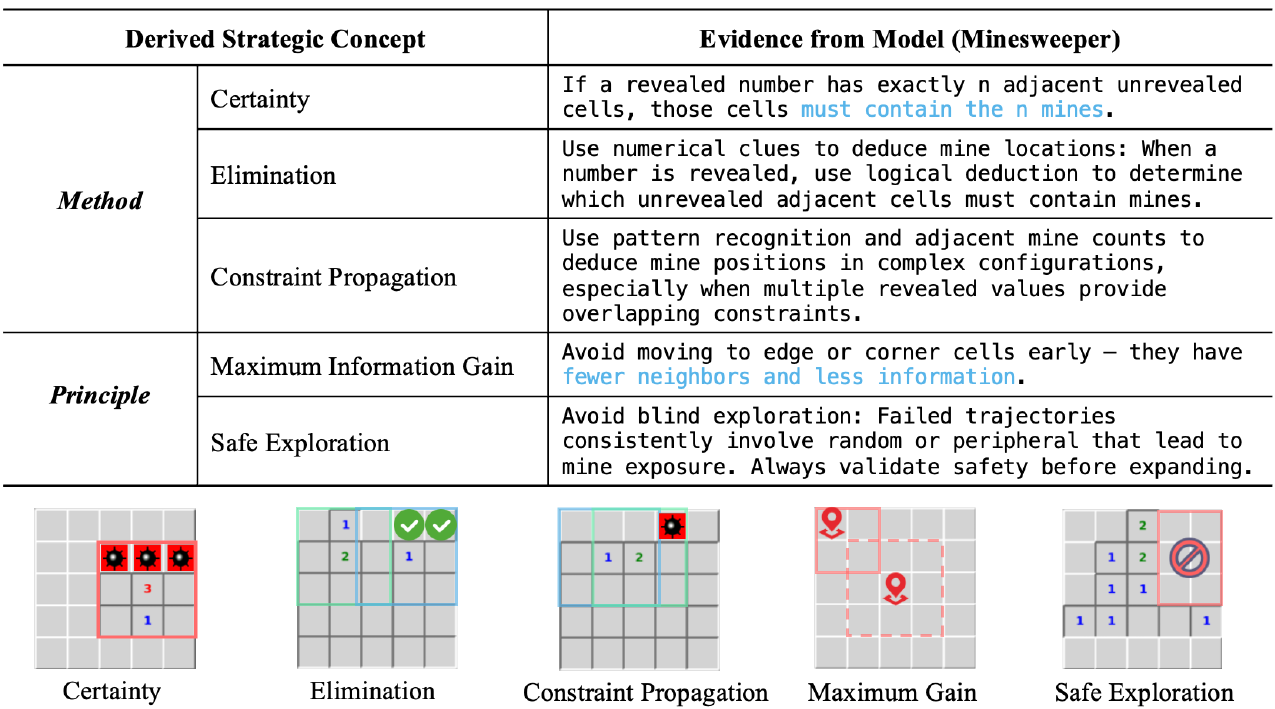}
    \caption{A synthesis of the strategic playbook $\Pi_k$ for Minesweeper, generated via Strategy and Playbook Summarization. The agent distills both tactical \textit{Methods} and high-level \textit{Principles} from its gameplay experience. This explicit playbook is used to condition the agent's value judgments, enabling more strategically sophisticated decision-making.}
    \label{fig:strategy}
\end{figure}

\subsection{Generalization}

To validate that CEL agent learns by understanding rather than memorization, we tested its generalization capabilities in two settings, summarized in Table~\ref{tab:generalization_results}.

First, for \textbf{intra-game generalization}, we evaluated the agent on 32 new seeds that were entirely unseen layouts from those used during training. The agent maintained a high level of performance on these unseen 256 instances, confirming that it learns the game's fundamental principles rather than overfitting to the specific training levels.
This highlights a fundamental paradigm difference from conventional reinforcement learning, which is notoriously hard to generalize to unseen domains. Instead of overfitting to learned patterns, the CEL agent's success on new layouts stems from its ability to apply an understanding of the game's rules to reason and plan effectively.

Furthermore, in the more challenging \textbf{inter-game generalization} setting, detailed in Figure~\ref{fig:inter_game}, we tested a model trained on one game in the environment of another.
Specifically, a Minesweeper-trained agent tested on Frozen Lake (left) and a FrozenLake-trained agent on Minesweeper (right) both show robust learning curves, despite their core model weights being frozen.
This success indicates that the agent transfers not its knowledge of game-specific rules, but rather its fundamental ability to learn by reasoning and planning when faced and interact with a novel environment. The CEL agent thus demonstrates a sophisticated ability to generalize not the concrete dynamics of a game, but the abstract wisdom of how to \emph{reason, plan} and then \emph{act}.

\begin{table}[!t]
\vspace{-8pt}
\centering
\caption{
The CEL agent's generalization performance across intra-game (unseen layouts) and inter-game (new game environments) settings.
The results demonstrate the agent's strong generalization, showcasing both robust performance on unseen layouts (intra-game) and successful zero-shot transfer to novel environments (inter-game).
Values in (.) show gain over the Zero-shot baseline, while those in [.] show change relative to the corresponding in-domain performance.
}
\begin{tabular}{c|cl|c}
\hline
\multirow{2}{*}{Trained On} & \multicolumn{2}{c|}{Tested On (Inter-Game)}                        & \multirow{2}{*}{\begin{tabular}[c]{@{}c@{}}Intra-Game\\ (Unseen layouts)\end{tabular}} \\ \cline{2-3}
                            & Minesweeper                      & \multicolumn{1}{c|}{FrozenLake} &                                                                                           \\ \hline
Zero-shot w/ Rule           & 25.8                             & \multicolumn{1}{c|}{78.9}       & -                                                                                         \\ \hline
Minesweeper                 & \multicolumn{1}{l}{53.5 (+27.7)} & 97.3 (+18.4)                    & 50.4 {[}-3.1{]}                                                                           \\
FrozenLake                  & \multicolumn{1}{l}{46.9 (+21.1)}  & 97.3 (+18.4)                    & 93.8 {[}-3.5{]}                                                                           \\ \hline
\end{tabular}
\label{tab:generalization_results}
\vspace{-8pt}
\end{table}

\begin{figure}[t!]
    \centering
    \includegraphics[width=\textwidth]{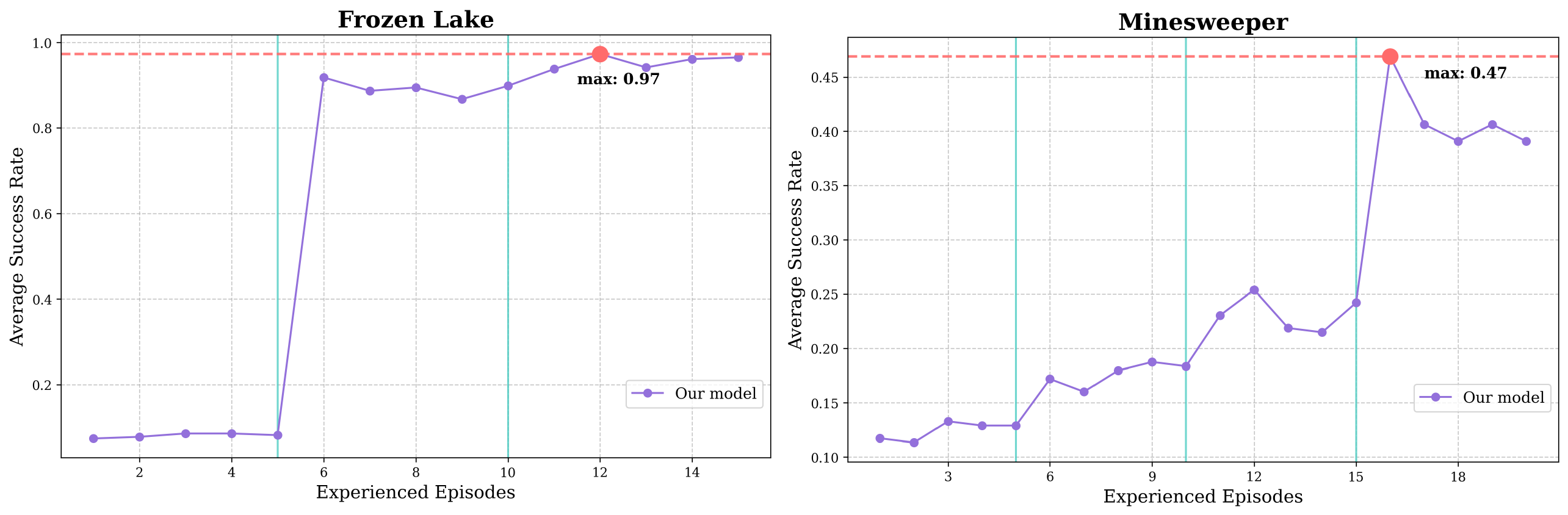}
    \caption{Inter-game generalization study showcasing adaptation to novel environments without model retraining. The plots show the agent's cross-game performance: a Minesweeper-trained agent on Frozen Lake (left) and a FrozenLake-trained agent on Minesweeper (right). In both evaluations, the core model weights remain frozen. The agent’s adaptation relies solely on the iterative refinement of its explicit rulebook and strategic playbook every 5 episodes (indicated by cyan lines).}
    \label{fig:inter_game}
\end{figure}

\end{section}

\begin{section}{Conclusions}\label{sec:conclusion}
In this work, we introduced \textit{Cogito, ergo ludo} (CEL), a novel agent architecture that learns by explicitly reasoning about its environment. Through a unique two-phase cycle of in-episode planning and post-episode reflection, CEL autonomously constructs a human-readable world model and strategic playbook from raw interaction, starting from a tabula rasa state. Our results across several environments demonstrate that this ``learning by thinking'' approach allows the agent to master complex tasks while creating a transparent and auditable decision-making process. CEL marks a significant departure from opaque, brute-force learning paradigms. It validates language-based reasoning as a powerful foundation for building agents that are not only capable but also interpretable and trustworthy, opening compelling pathways toward hybrid systems that fuse CEL's explicit understanding with traditional architectural efficiency.

\end{section}

\bibliography{iclr/iclr2026_conference}

\newpage
\appendix
\onecolumn
\newpage

This appendix extends upon the main paper by providing additional details on our experimental setup, ablation studies, qualitative results, and implementation specifics.
\begin{itemize}
    \item \textbf{Section~\ref{sec:appendix_env}:} Provides descriptions of the game environments used in our experiments.
    \item \textbf{Section~\ref{sec:ablation_action_only}:} Presents an analysis on the ``action-only'' model.
    \item \textbf{Section~\ref{sec:expanded_training}:} Shows the results of an additional experiment on Minesweeper with an expanded training set of 128 seeds, highlighting the scalability of our agent.
    \item \textbf{Section~\ref{sec:prompt_templates}:} Contains the full prompt templates used for the agent's in-episode decision-making and post-episode reflection.
    \item \textbf{Section~\ref{sec:additional_results}:} Provides concrete, qualitative examples of the agent's learned knowledge, including decision-making traces, environmental rules, and strategic playbooks for various games.
\end{itemize}

\section{Details of Environments}
\label{sec:appendix_env}

All game environments used in our experiments are from the TextArena~\citep{guertler2025textarena}. Below are the descriptions for the specific environments and configurations used in this work.

\textbf{Minesweeper} is a logic puzzle where the objective is to clear a grid of all non-mine cells without detonating any mines. When a cell is revealed, it displays a number indicating how many adjacent cells contain mines, and the player must use this information to deduce the location of the mines. In our experiments, the game is configured on a 5$\times$5 grid with 3 randomly placed mines.

\textbf{Frozen Lake} is a canonical grid navigation problem on a 6$\times$6 grid where an agent must travel from a start tile to a goal tile, avoiding 6 randomly placed holes. In our deterministic setting, each action moves the agent exactly one cell in the chosen direction, removing the stochastic ``slippery'' nature often associated with this environment. This modification allows for a direct assessment of the agent's planning and rule-induction capabilities without the confounding factor of environmental randomness.

\textbf{Sokoban} is a classic puzzle game where the player must push all boxes to designated goal locations. The player can only push one box at a time and cannot pull boxes. This simple constraint creates a complex search space and necessitates careful, long-horizon planning to avoid irreversible states, such as trapping a box in a corner. The version used in our study is played on a 6$\times$6 grid with a single box.

\section{Analysis on Rollout Outcomes for Action-only Model}
\label{sec:ablation_action_only}

\begin{figure}[!h]
    \centering
    \includegraphics[width=0.5\textwidth]{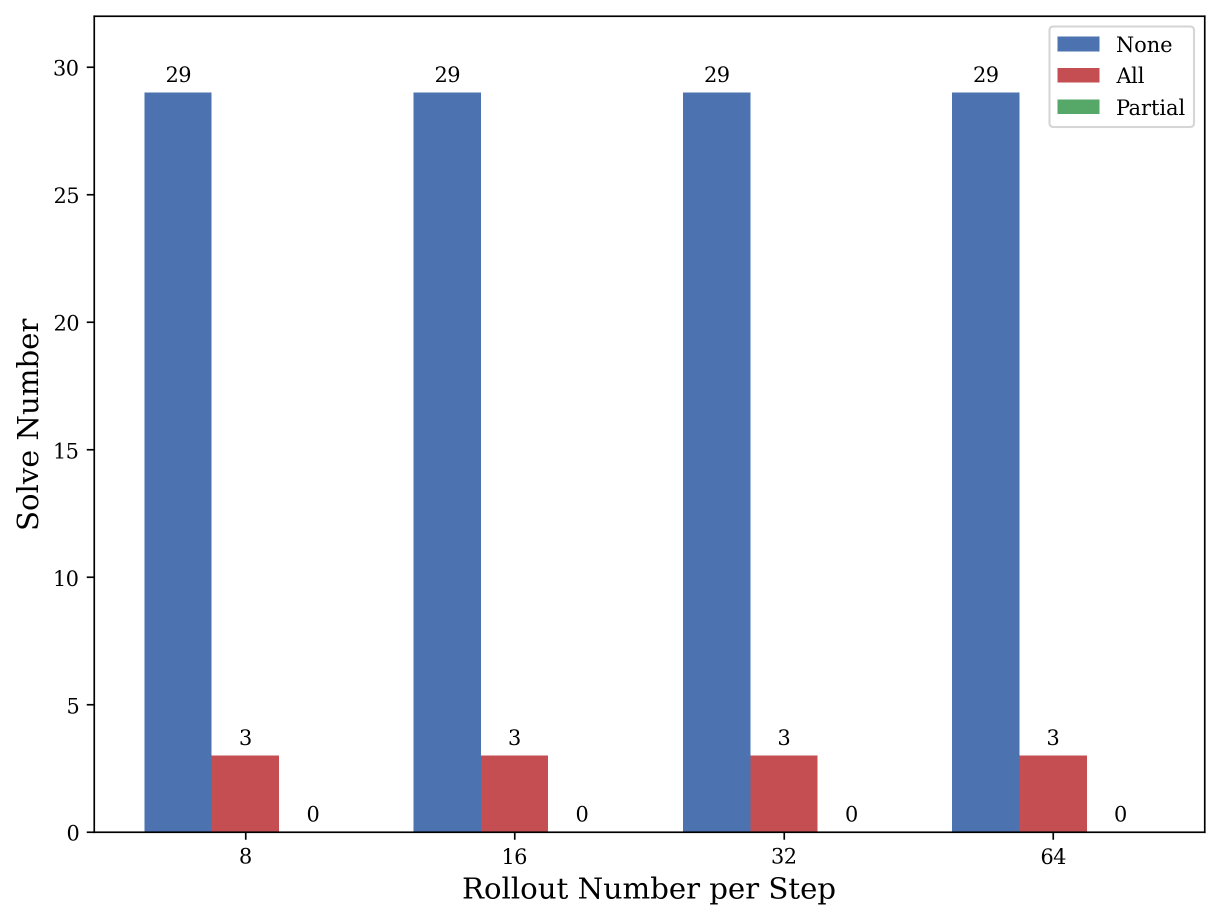}
    \caption{Distribution of rollout outcomes for the Action-only model. Across all group sampling sizes (8 to 64), outcomes are polarized into ``All'' (all succeed) or ``None'' (all fail). The number of ``Partial'' rollouts, which are required for GRPO to learn, is consistently zero, causing training failure.}
    \label{fig:rollout_distribution}
\end{figure}

To understand the necessity of cognitive components and chain-of-thought reasoning, we ablated it with a simplified ``Action-only'' agent that directly outputs actions. When attempting to train this agent with GRPO, we observed a consistent training failure. The reason lies in the extreme polarization of rollout outcomes, as shown in Figure~\ref{fig:rollout_distribution}. GRPO requires batches with mixed results (i.e., ``partially successful'') to derive a learning signal. However, across all tested sampling sizes (8 to 64), the outcomes for the Action-only agent were always binary: for any given seed, all rollouts in a batch either succeeded or failed. As a result, the number of partially successful rollouts, i.e., the sole source of a viable training signal, was consistently zero. This lack of comparative data within batches starves the GRPO algorithm of a gradient, leading to a breakdown in training and highlighting the critical role of nuanced reasoning traces in enabling effective optimization.

\section{Performance with an Expanded Training Set}
\label{sec:expanded_training}

To further evaluate the scalability and generalization capabilities of our CEL agent, we conducted an additional experiment on the Minesweeper environment. We expanded the set of training layouts by increasing the number of unique seeds from 32 (used in the main experiments) to 128. 

\begin{figure}[!h]
    \centering
    \includegraphics[width=0.5\textwidth]{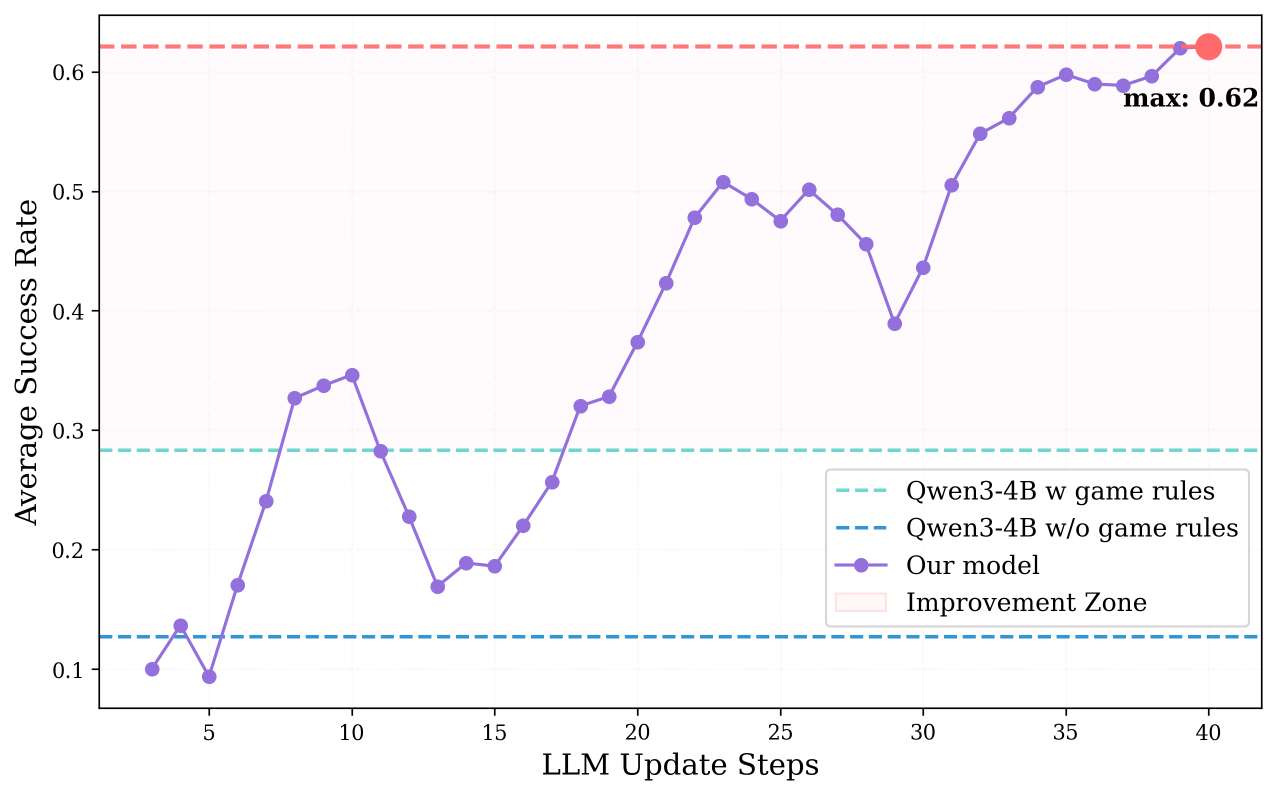}
    \caption{Learning curve for the CEL agent on Minesweeper when trained on an expanded set of 128 unique seeds. The agent achieves a new peak success rate of 62\%, surpassing the performance observed with 32 seeds.}
    \label{fig:results_minesweeper_128}
\end{figure}

The results are presented in Figure \ref{fig:results_minesweeper_128}. The agent demonstrates a notable improvement in performance, with its peak success rate climbing from 54\% (as reported in Figure~\ref{fig:train_curves}) to a new maximum of 62\%. This finding suggests that exposure to a more diverse set of game scenarios directly enhances the agent's core reasoning and planning capabilities. This confirms that the agent is developing a robust, generalizable problem-solving model for the game, rather than overfitting to a limited number of specific layouts.

\section{Prompt Templates}
\label{sec:prompt_templates}

\begin{figure}[!h]
    \centering
    \includegraphics[width=\textwidth]{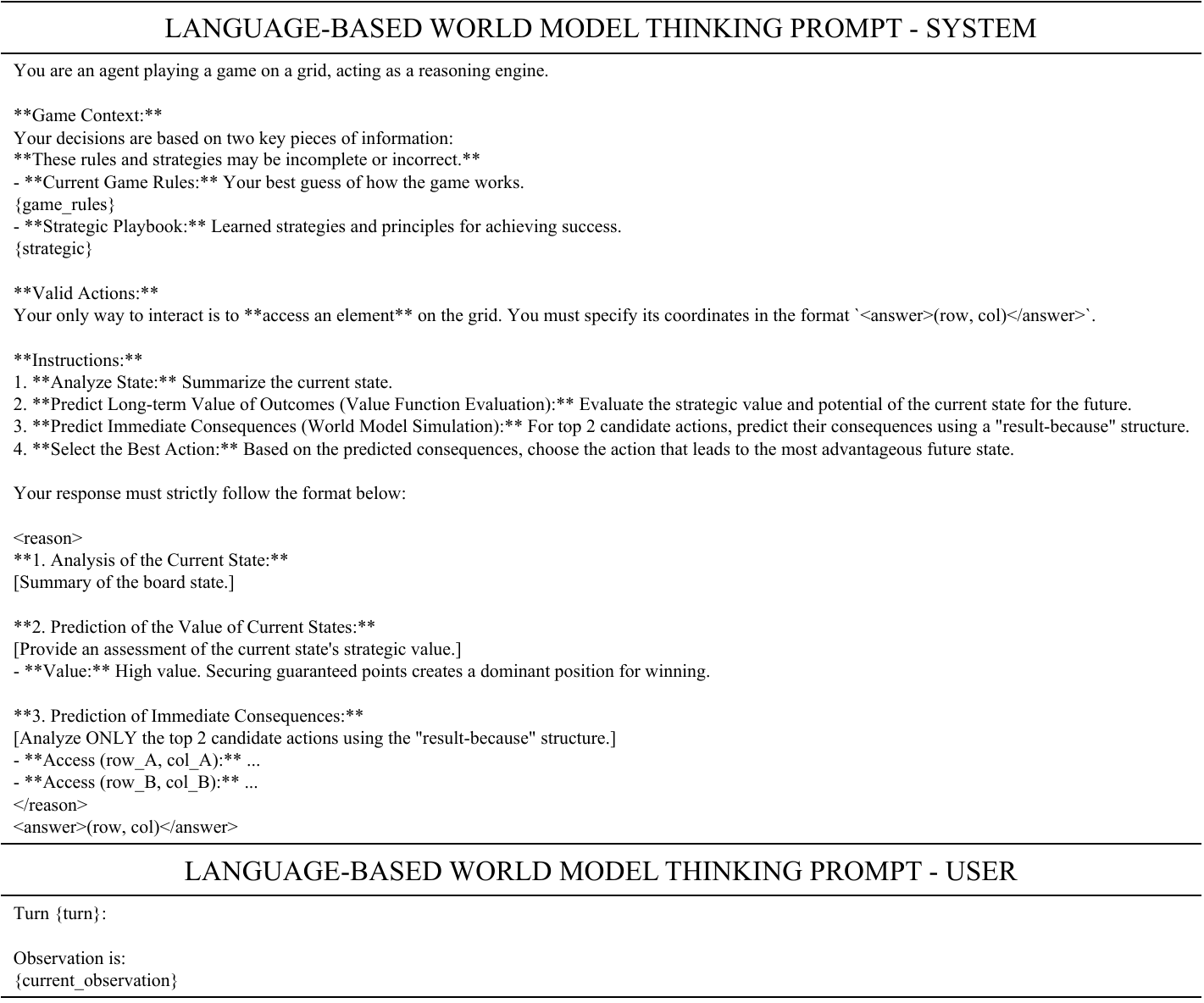}
    \caption{The prompt template for in-episode decision-making (Phase 1). It instructs the LLM to evaluate the current state, assess their strategic value (LVF) and predict action outcomes (LWM).}
    \label{fig:prompt_thinking}
\end{figure}

\begin{figure}[!h]
    \centering
    \includegraphics[width=\textwidth]{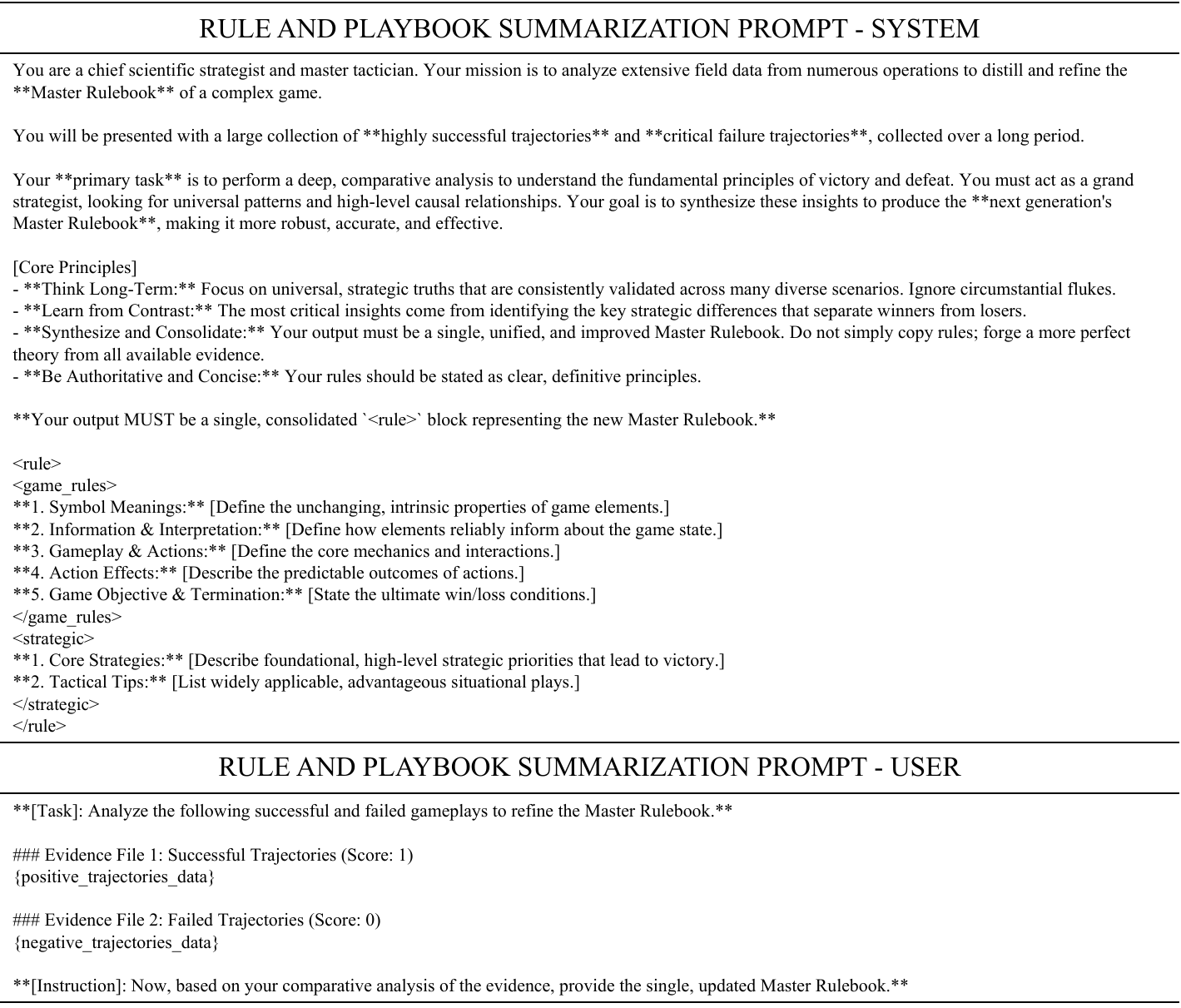}
    \caption{The prompt template for the Rule Induction and Playbook Summarization process (Phase 2). It guides the LLM to analyze a completed episode's trajectory and refine its explicit model of the environment's dynamics.}
    \label{fig:prompt_rule}
\end{figure}

In this section, we present the core prompt templates used by the CEL agent. Figure \ref{fig:prompt_thinking} shows the template for in-episode decision-making, and Figure \ref{fig:prompt_rule} shows the template for post-episode reflection.

\section{Additional Results}
\label{sec:additional_results}

\begin{figure}[h!]
    \centering
    \includegraphics[width=1\textwidth]{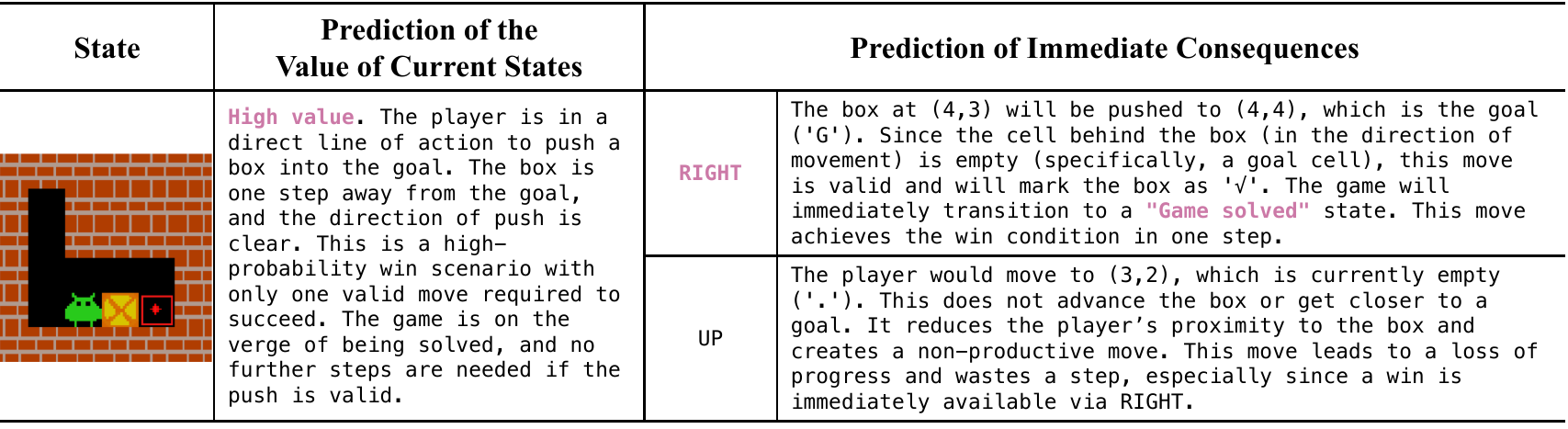}
    \caption{An example of a Decision-Making process for Sokoban environment.
    }
    \label{fig:decision_making}
\end{figure}

\begin{figure}[h!]
    \centering
    \includegraphics[width=1\textwidth]{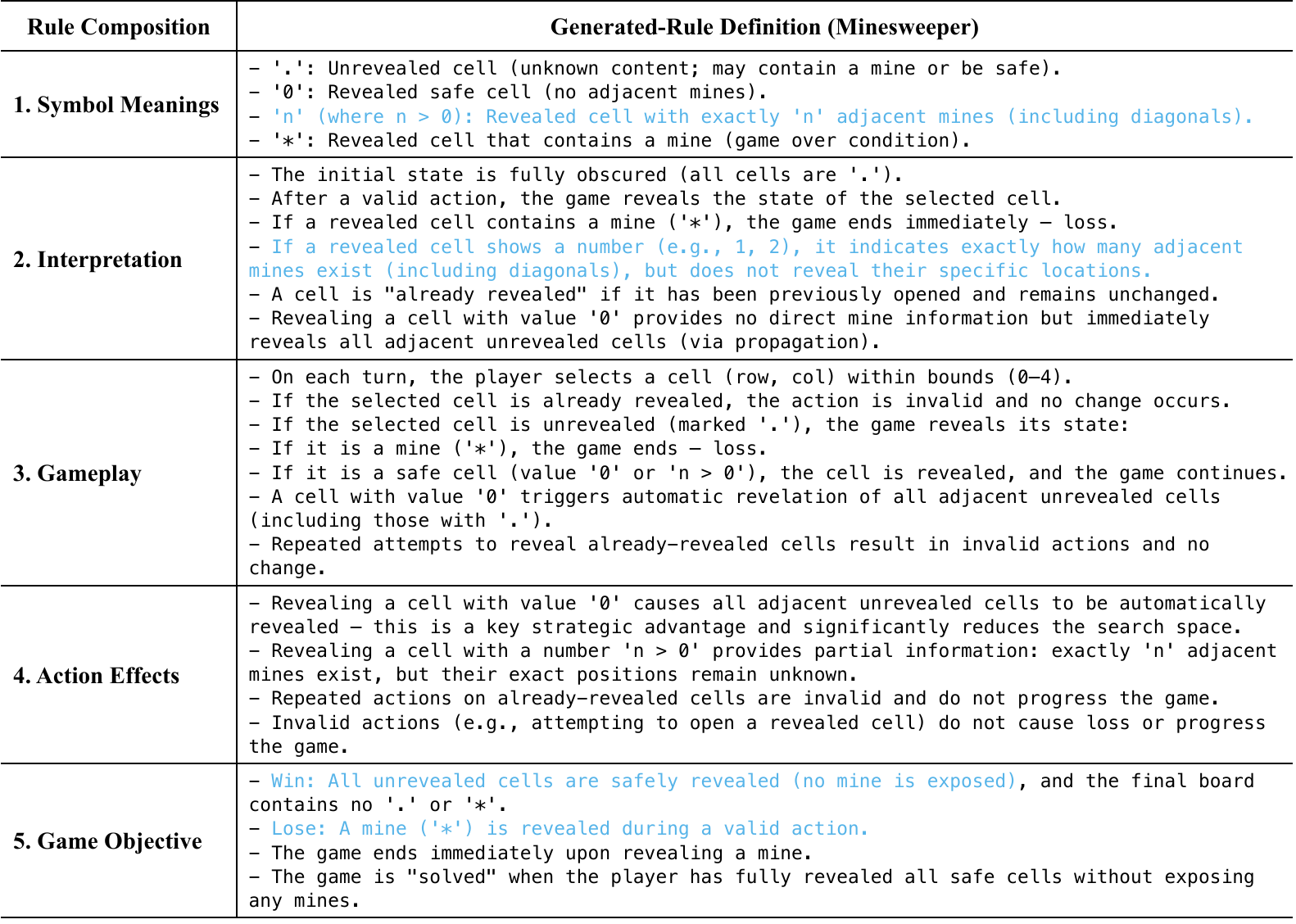}
    \caption{An example of a learned environmental rule for Minesweeper environment.
    }
    \label{fig:minesweeper_rule_full}
\end{figure}

\begin{figure}[!h]
    \centering
    \includegraphics[width=1\textwidth]{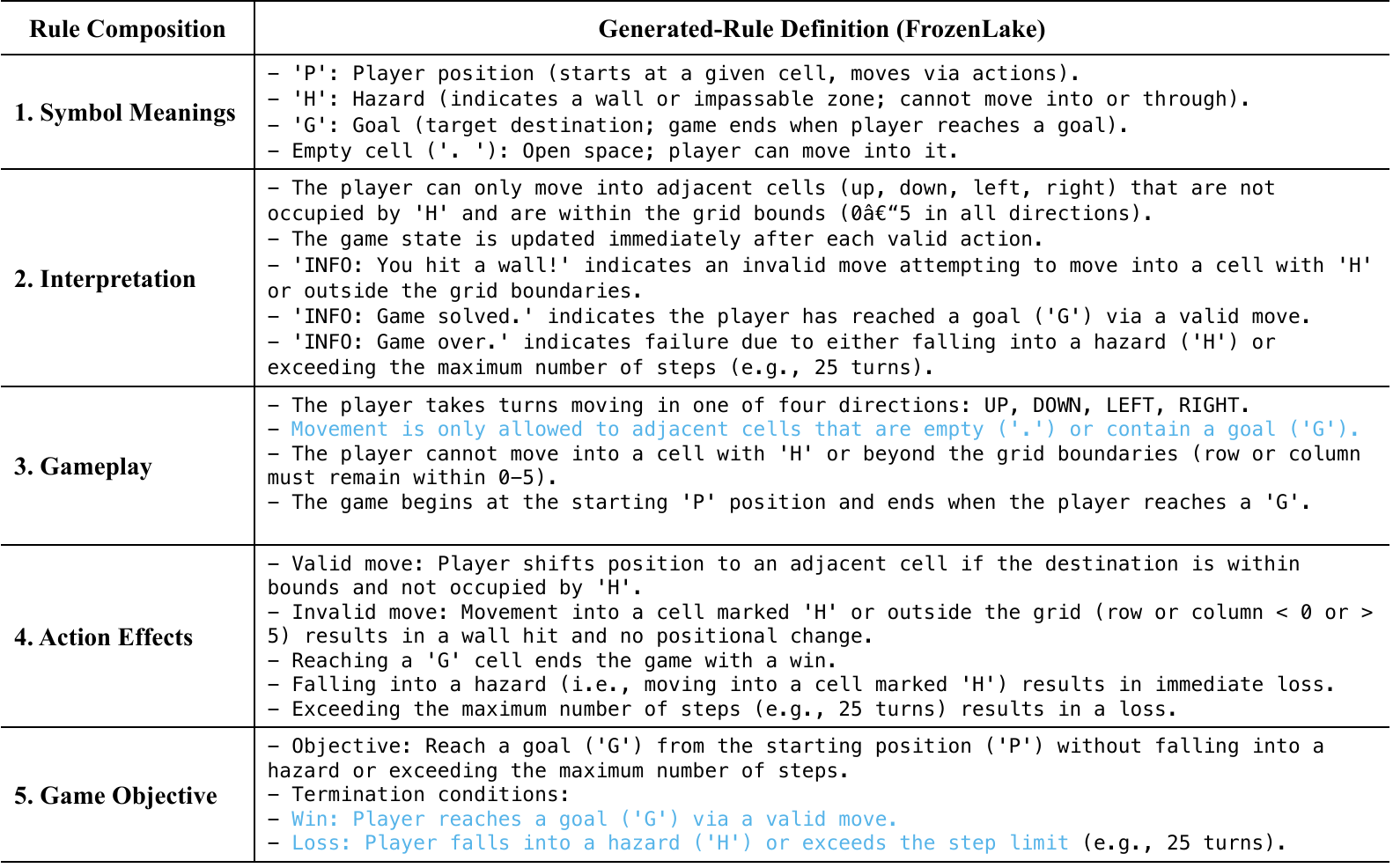}
    \caption{An example of a learned environmental rule for FrozenLake environment.}
    \label{fig:frozenlake_rule}
\end{figure}

\begin{figure}[!h]
    \centering
    \includegraphics[width=1\textwidth]{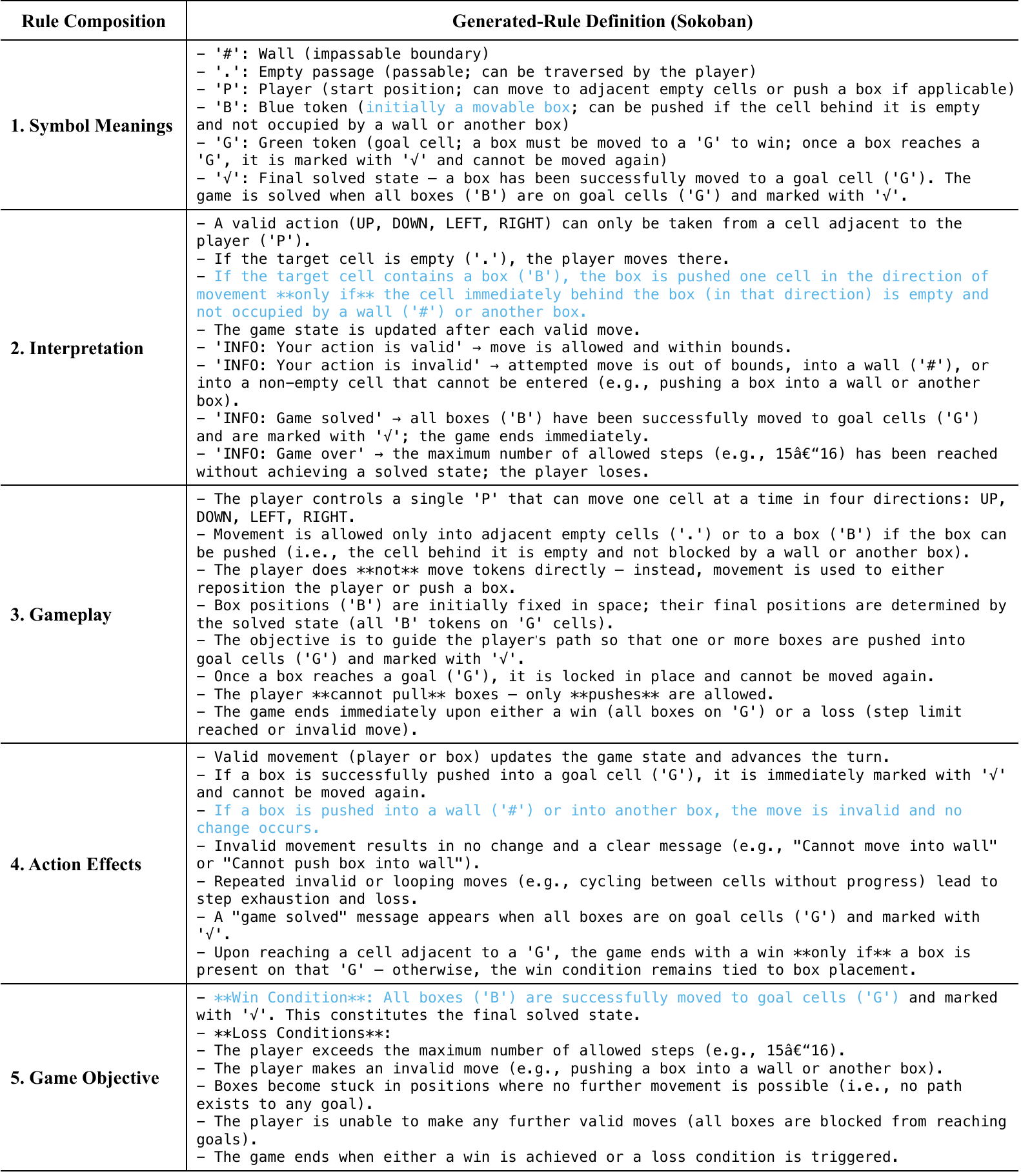}
    \caption{An example of a learned environmental rule for Sokoban environment.}
    \label{fig:sokoban_rule}
\end{figure}

\begin{figure}[!h]
    \centering
    \includegraphics[width=1\textwidth]{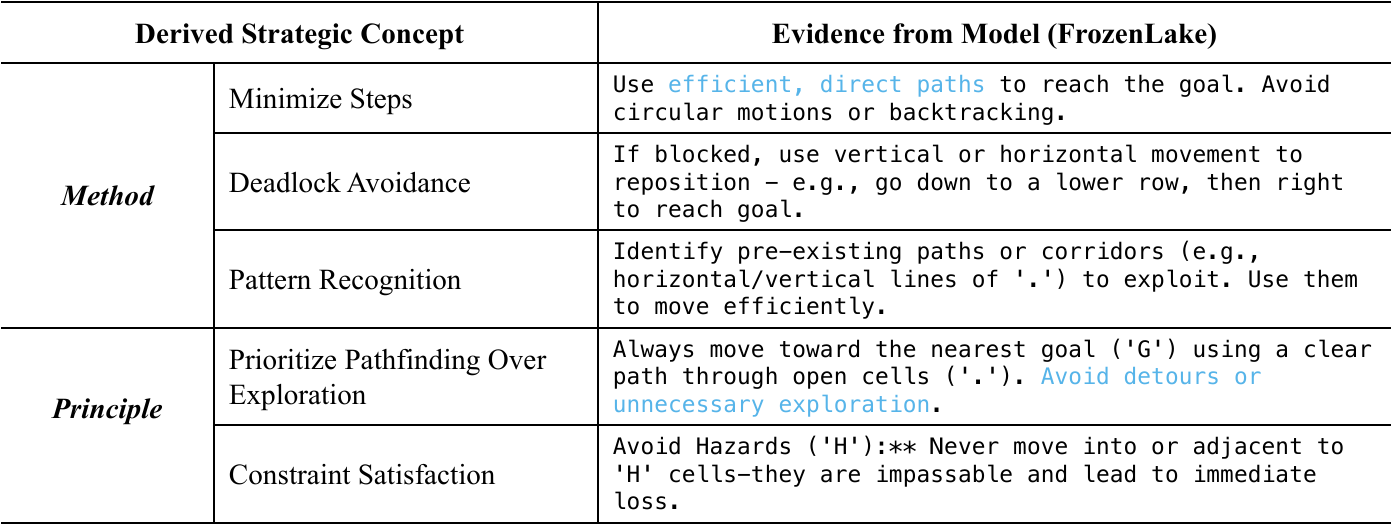}
    \caption{An example of a learned strategic guideline from the agent's playbook for FrozenLake environment.}
    \label{fig:frozenlake_strategy}
\end{figure}

\begin{figure}[!h]
    \centering
    \includegraphics[width=1\textwidth]{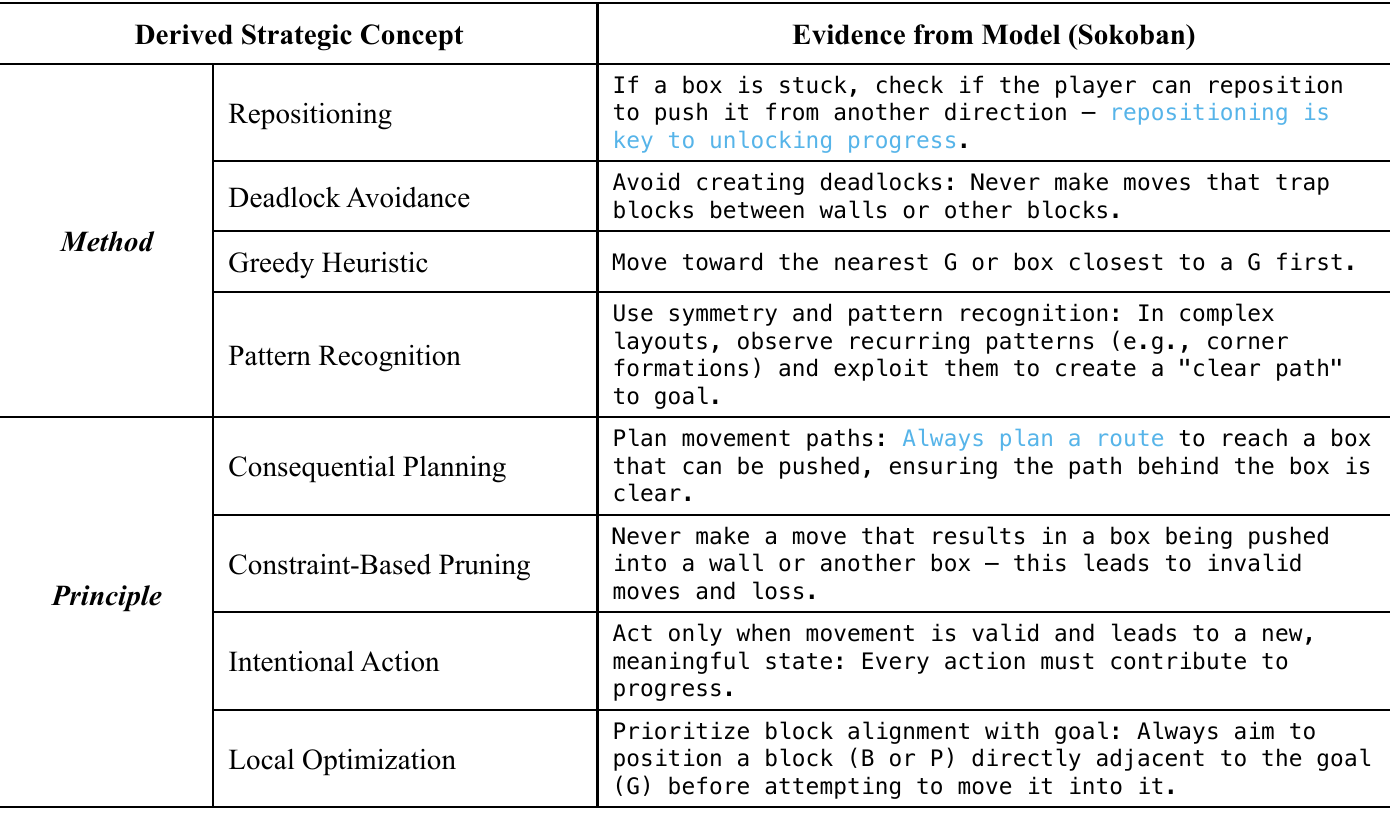}
    \caption{An example of a learned strategic guideline from the agent's playbook for Sokoban environment.}
    \label{fig:sokoban_strategy}
\end{figure}

To illustrate the explicit and interpretable knowledge base generated by our CEL agent, we provide concrete examples of Decision-Making processes (Figure~\ref{fig:decision_making}), learned environmental rules (Figure~\ref{fig:minesweeper_rule_full}, Figure~\ref{fig:frozenlake_rule}, Figure~\ref{fig:sokoban_rule}) and strategic playbooks (Figure~\ref{fig:frozenlake_strategy}, Figure~\ref{fig:sokoban_strategy}) for the Minesweeper, FrozenLake and Sokoban environment.

\end{document}